\pdfoutput=1

\documentclass[11pt]{article}

\usepackage{coling}

\usepackage{times}
\usepackage{latexsym}

\usepackage[T1]{fontenc}

\usepackage[utf8]{inputenc}

\usepackage{microtype}

\usepackage{inconsolata}

\usepackage{graphicx}

\usepackage{amsmath}
\usepackage{microtype}
\usepackage{graphicx}
\usepackage{subfigure}
\usepackage{booktabs} 
\usepackage{multicol}
\usepackage{courier} 
\usepackage{listings, xcolor}
\usepackage{dirtree}
\usepackage{hyperref}
\usepackage{dirtree}
\usepackage{titletoc}

\newcommand{\devterm}{DevEval}

\usepackage{dcolumn} 

\newcolumntype{d}[1]{D{.}{.}{#1}}
\usepackage{tikz}
\usetikzlibrary{shapes.geometric}

\newcommand{\mytag}[2]{%
  \tikz[baseline=(X.base)]
    \node [draw=#2,
           fill=#2,
           text=black,
           rounded corners,
           text height=1.5ex,
           text depth=.25ex] (X) {#1};
}
\definecolor{c1}{RGB}{255, 220, 220}
\definecolor{c2}{RGB}{255, 255, 204}
\definecolor{c3}{RGB}{229, 255, 204}
\definecolor{c4}{RGB}{220, 255, 255}
\definecolor{c5}{RGB}{220, 220, 255}
\definecolor{c6}{RGB}{230, 230, 230}
\definecolor{c7}{RGB}{255, 229, 204}
\definecolor{c8}{RGB}{249, 224, 255}

\newcommand{\opt}[1]{{\fontfamily{lmtt}\selectfont #1}}

\title{Prompting Large Language Models to Tackle \\the Full Software Development Lifecycle: A Case Study}

\author{
 \textbf{Bowen Li\textsuperscript{1}\thanks{Equal contribution.}},
 \textbf{Wenhan Wu\textsuperscript{2}\footnotemark[1]\thanks{Work done during internship at Shanghai AI Laboratory.}},
 \textbf{Ziwei Tang\textsuperscript{3}\footnotemark[1]\footnotemark[2]},
 \textbf{Lin Shi\textsuperscript{4}\footnotemark[1]\footnotemark[2]},
 \textbf{John Yang\textsuperscript{5}},
 \textbf{Jinyang Li\textsuperscript{6}},
\\
 \textbf{Shunyu Yao\textsuperscript{5}},
 \textbf{Chen Qian\textsuperscript{7}},
 \textbf{Binyuan Hui\textsuperscript{8}},
 \textbf{Qicheng Zhang\textsuperscript{1}},
 \textbf{Zhiyin Yu\textsuperscript{1}},
 \textbf{He Du\textsuperscript{1}},
\\
 \textbf{Ping Yang\textsuperscript{9}},
 \textbf{Dahua Lin\textsuperscript{1}},
 \textbf{Chao Peng\textsuperscript{9}\thanks{Corresponding authors.}},
 \textbf{Kai Chen\textsuperscript{1}\footnotemark[3]}
\\
\\
 \textsuperscript{1}Shanghai AI Laboratory,
 \textsuperscript{2}Nanjing University,
 \textsuperscript{3}BUPT,
\\
 \textsuperscript{4}Dartmouth College,
 \textsuperscript{5}Princeton University,
 \textsuperscript{6}The University of Hong Kong,
\\
 \textsuperscript{7}Tsinghua University,
 \textsuperscript{8}Alibaba Group,
 \textsuperscript{9}ByteDance
\\
 \small{
   \textbf{Correspondence:} \href{mailto:chenkai@pjlab.org.cn}{chenkai@pjlab.org.cn} and \href{mailto:pengchao.x@bytedance.com}{pengchao.x@bytedance.com}
 }
}

\begin{document}

\maketitle
\begin{abstract}
Recent advancements in large language models (LLMs) have significantly enhanced their coding capabilities.
However, existing benchmarks predominantly focused on simplified or isolated aspects of coding, such as single-file code generation or repository issue debugging, falling short of measuring the full spectrum of challenges raised by real-world programming activities. 
In this case study, we explore the performance of LLMs across the entire software development lifecycle with \devterm, encompassing stages including software design, environment setup, implementation, acceptance testing, and unit testing.
\devterm~features four programming languages, multiple domains, high-quality data collection, and carefully designed and verified metrics for each task. 
Empirical studies show that current LLMs, including GPT-4, fail to solve the challenges presented within \devterm. 
Our findings offer actionable insights for the future development of LLMs toward real-world programming applications.
\footnote{Our data and code are available at \url{https://github.com/open-compass/DevEval}.}
\end{abstract}

\section{Introduction}
\begin{figure*}[tb]
  \centering
  \includegraphics[width=\textwidth]{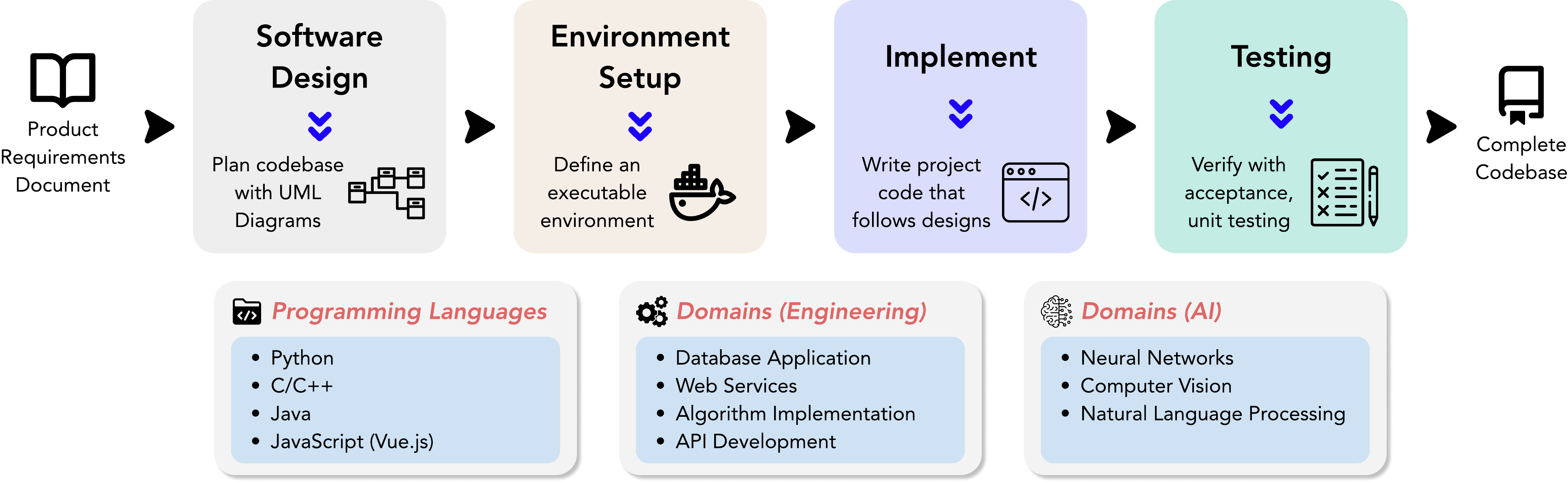}
  \caption{Our \devterm~features multiple stages of software development, including software design, environment setup, implementation, and testing (both acceptance and
unit testing).}
  \label{fig:intro}
\end{figure*}

\begin{table*}
    \centering
     \scalebox{0.75}{
    \begin{tabular}{l|cccc}
    \toprule
    \textbf{Task} & \textbf{Input$^{\dagger}$} & \textbf{Output}                                              & \textbf{Environment} & \textbf{Evaluation}   \\
    \midrule
    Software Design     & PRD                                                                                                                   & \begin{tabular}[c]{@{}l@{}}UML Diagrams$^\ddag$, \\ Architecture Design\end{tabular} & N/A                     & Subjective Evaluation       \\
    \midrule
    Environment Setup  & \begin{tabular}[c]{@{}l@{}}PRD, UML Diagrams, \\ Architecture Design\end{tabular}                                     & Dependency Files                                                              & Base      & Pass Rate on Usage Examples \\
    \midrule
    Implementation     & \begin{tabular}[c]{@{}l@{}}PRD, UML Diagrams, \\ Architecture Design\end{tabular}                                     & Implementation Code                                                          & Reference & Unit \& Acceptance Testing  \\
    \midrule
    Acceptance Testing & \begin{tabular}[c]{@{}l@{}}PRD, UML Diagrams, \\ Architecture Design,  \\ Implementation Code$^{\ast}$ \end{tabular} & Acceptance Testing Code                                                      & Reference & Oracle Test                 \\
    \midrule
    Unit Testing       & \begin{tabular}[c]{@{}l@{}}PRD, UML Diagrams, \\ Architecture Design,  \\ Implementation Code \end{tabular} & Unit Testing Code                                                            & Reference & Oracle Test \& Coverage    \\
    \bottomrule
    \end{tabular}
    }
    \caption{Task design in DevEval. ${\dagger}$: following our modular evaluation protocol, the input for each task are reference. 
    ${\ddagger}$: includes UML class and sequence diagrams.
    ${\ast}$: implementation source code is optional for acceptance testing.}
    \label{tab:tasks}

\end{table*}

Given its practical value and reasoning challenges, programming has become an important domain to deploy and evaluate large language models (LLMs), leading to popular products like GitHub Copilot
and benchmarks like HumanEval~\citep{chen2021evaluating} and APPS~\citep{hendrycksapps2021}. 
While these earlier coding tasks focused on generating a single code file or even a single method from simple instructions, recent works such as SWE-bench~\cite{jimenez2023swe} and RepoBench~\cite{liu2023repobench} evaluate LLMs on repository-level tasks, which feature longer, more involved NL2Code problems.
Still, these benchmarks concentrate on narrow aspects of software development, leaving a gap in comprehensive studies that encompass the full software development lifecycle across its various phases.

To address these shortcomings and fill this gap, we present \devterm, a comprehensive case study that mirrors real-world software development.
\devterm~generally evaluates models on the task of constructing a multi-file codebase starting from a product requirement document (PRD) of detailed specifications.
Subscribing to the traditional Waterfall software development model~\cite{royce1987managing}, \devterm~breaks down this process into a diverse set of inter-related development stages, i.e., software design, environment setup, implementation, acceptance and unit testing, as visualized in Figure~\ref{fig:intro} and Table~\ref{tab:tasks}.
In contrast to previous works, \devterm~is the first to evaluate models' software design and environment setup capabilities.
One significant challenge in this study lies in the scarcity of publicly available repositories that include the full range of software development artifacts, particularly design documents and comprehensive testing programs.
To overcome this, we curated a collection of 22 repositories across four programming languages (Python, C/C++, Java, JavaScript), spanning various domains such as machine learning, web services, and command-line utilities.
By encompassing the multi-faceted, interconnected steps of software development under a single framework, \devterm~provides a holistic view of LLMs' capabilities for automated software production, moving beyond the conventional focus on code completion.

Through a comprehensive experimental study, our findings indicate that tested models struggle significantly with the challenges presented.
GPT-4-Turbo achieves the highest scores amongst all evaluated models, yet it obtains less than 10\% on our repository-level implementation task.
Other tasks prove relatively more manageable for models; however, even GPT-4-Turbo struggles to attain scores exceeding 40\% on the more complicated ones. For instance, models generally fail to generate executable tests, with oracle test scores falling below 40\%.
Despite this, the generated testing code demonstrates potential in code coverage, achieving as high as 79.4\% when it is executable.
Furthermore, our investigation into different prompting methods shows that prompts  with external information and execution feedback could yield notable and consistent improvements.
Importantly, qualitative analysis shows that models demonstrate difficulties in handling Makefile and Gradle, configuring function arguments and employing advanced object-oriented programming techniques.
Overall, \devterm~introduces a novel challenge for existing LLMs, and our investigation sheds light on fundamental issues, paving the way for future research.

\section{Related Work}

Initial code generation datasets focused on self-contained, Python-based completion problems~\cite{chen2021evaluating,hendrycksapps2021,austin2021program}, while later works expanded complexity by increasing language coverage~\cite{zheng2023codegeex,Cassano2023MultiPLEAS}, enhancing execution-based test coverage~\cite{evalplus,Wang2022ExecutionBasedEF}, incorporating dependencies~\cite{Lai2022DS1000AN,Ding2023CrossCodeEvalAD,Liu2023MLBenchLL}, introducing interactive environments~\cite{Yin2022NaturalLT,Yang2023InterCodeSA}, and developing short-form task suites~\cite{lu2021codexglue,Muennighoff2023OctoPackIT}. \devterm~ aligns with recent repository-scale coding works~\cite{jimenez2023swe,liu2023repobench,Zhang2023RepoCoderRC}.
However, our work distinctly evaluates LLMs' ability to create entire codebases from extensive natural language descriptions, offering a more comprehensive test of LLM capabilities. 
In automatic programming, role-play frameworks with communicative agents~\cite{hong2023metagpt, li2023camel, qian2023communicative, qian2023experiential} like MetaGPT~\cite{hong2023metagpt} integrate structured workflows and task modularity, similar to our approach. 
Compare against MetaGPT, \devterm~stands out by providing a structured evaluation across all phases of the software development lifecycle. 
Additionally, there is a growing trend in agent-based frameworks specially targeting software engineering tasks~\cite{yang2024sweagent, zhang2024autocoderover, liu2024marscodeagentainativeautomated, wang2024opendevinopenplatformai}.

\section{\devterm}

In this section, we discuss the design of \devterm, including task specifications and the evaluation criteria (summarized in Table~\ref{tab:tasks}).
Adhering to the modular approach, our design utilizes reference inputs for each task.
This strategy enables and concentrates on evaluating the efficacy of models in executing specific tasks.
\footnote{
Notably, our framework can be adeptly configured to facilitate end-to-end evaluations, utilizing the intermediate outputs generated by models across multiple tasks. 
Moreover, \devterm~ can also function in a Copilot mode, empowering human users to intervene and refine model outputs, thus enhancing the collaborative synergy between human expertise and automated systems.
}

\subsection{Task 1: Software Design}
\label{sec:design}
During this phase, the model is tasked with interpreting the PRD to create the design of the system before development. The first subtask involves the generation of the UML class diagram using Mermaid syntax, which models the structural aspects of software systems, detailing the classes within the system, their attributes, operations and the relationship among them. Class diagrams help developers understand the system's foundation before development begins. The second subtask focuses on the creation of the UML sequence diagram using Mermaid syntax. These diagrams clarify the collaboration among components of the system by mapping out the interaction between objects and processes over time, illustrating the sequence of messages exchanged between objects to implement the system's functionality. The final subtask fomulates architectural designs using hierarchical file tree structures, aiming to establish a structured framework for the source code, build scripts and necessary auxiliary files. The architectural design ensures a cohesive structure for coding, testing and maintenance.

\textbf{Evaluation.} 
Since the Software Design tasks are open-ended, we employ the \emph{LLM-as-a-judge} approach \citep{zheng2023judging, wang2023large, chiang2023can} to conduct the automatic evaluation.
The evaluation is anchored by two principal metrics: \emph{general principles} and \emph{faithfulness}. 

The \emph{general principles} metric plays a crucial role, with each task sharing common elements while maintaining specific criteria. 
For all the sub-tasks, principles like cohesion and decoupling, and practicability are fundamental. 
Cohesion and decoupling emphasize the importance of clarity and functionality within individual elements (classes or sequences) and reducing dependencies between different components. 
In terms of practicability, all tasks require designs to be readable, understandable, and modular, facilitating ease in development, testing, and maintenance. 
Meanwhile, each task has its unique focus areas: UML Class diagrams are evaluated on complexity;
UML Sequence diagrams concentrate on uniformity, integration, and interaction complexity;
Architecture Design highlights the distinction between design and coding, and conformance to practical standards.
Subsequently, the \emph{faithfulness} metric gauges the extent to which models adhere to specified instructions.

\subsection{Task 2: Environment Setup}

In the second phase of \devterm, models are provided with the PRD, UML diagrams, and architecture design to generate a dependency file for initializing the development environment.
This step is followed by the deployment of a standard installation command utilizing the generated file. For Python, the \texttt{Conda} environment manager is employed; for Java and JavaScript, \texttt{Gradle} and \texttt{NPM} are utilized respectively.\footnote{
It is noteworthy that our evaluation does not contain an Environment Setup task for C/C++ due to the absence of a universally acknowledged and user-friendly dependency management system for these languages~\cite{miranda2018use}.}
Setting up an environment often encounters challenges such as missing or outdated dependencies, along with version conflicts, all of which must be resolved to ensure a seamless development environment. 
Our research aims to investigate the potential of LLMs in automating this cumbersome process, thereby enhancing production efficiency.

\textbf{Evaluation.} The evaluation centers on the execution of dependency files across each programming language within a predetermined base environment delineated in a Docker file. This is followed by the execution of the repository's example usage code. The principal metric for evaluation in this task is the success rate of the executed example code.


\subsection{Task 3: Implementation}

For this task, models are provided with the PRD, UML diagrams and architecture design, and are then instructed to develop code for each source code file as specified in the architecture design.
Diverging from existing benchmarks on repository-level code generation \citep{liu2023repobench}, the implementation task in \devterm~ is dedicated to assessing LLMs in generating an entire code repository from scratch. A key innovation in our investigations is the detailed level of requirements provided to the LLMs. In contrast to similar systems like MetaGPT \citep{hong2023metagpt} and ChatDev \citep{qian2023communicative, qian2023experiential}, which generate outputs from brief requirement descriptions typically under 100 words, DevEval offers document-level detail to guide the models. This approach ensures the products are more precisely aligned with expectations, and further acceptance testing is employed for verification. As a result, our evaluations better reflect real-world software development scenarios where detailed requirements are essential to capture complex specifications and ensure product quality.


To more accurately simulate real-world development practices and ensure rigorous evaluation, the implementation task in \devterm~ involves supplying LLMs with comprehensive inputs including the PRD, UML class and sequence diagrams, and architecture design. 
The models are then prompted to generate code files. 
Given the constraint of output length, we adopt a sequential generation approach, prompting the models to produce one code file per interaction.
Regarding the planning aspect, recent studies have explored prompting LLMs or training a specific planner \citep{yao2023tree, besta2023graph, wang2023describe}. 
Considering the structured nature of code files and the inherent dependencies among them, we utilize these dependencies as a clue for effective planning. 
The generation process is guided to adhere to a partial order derived from a predefined directed acyclic graph, thereby ensuring structured and logical code development. 
We leave the exploration of planning generated by models themselves for future work.

\textbf{Evaluation.} For the evaluation of the implementation task, an automated testing framework has been developed.
This framework, tailored to the specific programming language in use, integrates \texttt{PyTest} for Python, \texttt{GTest} for C++, \texttt{JUnit} for Java, and \texttt{Jest} for JavaScript.
The evaluation procedure involves executing reference acceptance and unit tests within a predefined reference environment.  Then the evaluation metric is determined by the pass rate of these tests.

\begin{table}[ht]
\centering
\scalebox{0.6}{
\begin{tabular}{c|c|c|c|c}
\toprule
 & \textbf{Python} & \textbf{C/C++} & \textbf{Java} & \textbf{JavaScript$^\ddag$}  \\
\midrule
Domain$^{\dag}$ & \begin{tabular}[c]{@{}c@{}}\mytag{NLP}{c1}\mytag{CV}{c2}\\ \mytag{DL}{c3}\mytag{ALGO}{c4}\\ \mytag{API}{c5}\mytag{Tool}{c8}\end{tabular}  & \begin{tabular}[c]{@{}c@{}} \mytag{DB}{c6} \\ \mytag{ALGO}{c4}\\ \mytag{Tool}{c8}\end{tabular} & \begin{tabular}[c]{@{}c@{}}\mytag{CV}{c2}\mytag{DB}{c6} \\ \mytag{ALGO}{c4} \\ \mytag{Tool}{c8} \end{tabular}  & \mytag{WEB}{c7}            \\
\midrule
\#Repo                                                         & 10     & 5     & 5    & 2          \\
\midrule
\begin{tabular}[c]{@{}c@{}}Avg. \\ \#Code File\end{tabular}    & 2.2    & 7.0   & 5.4  & 6.0        \\
\midrule
\begin{tabular}[c]{@{}c@{}}Avg. \\ \#Code Line\end{tabular}    & 276    & 495   & 524  & 617       \\
\midrule
\begin{tabular}[c]{@{}c@{}}Avg. \\ \#Accep. Tests\end{tabular} & 3.0    & 5.4   & 2.4  & 2        \\
\midrule
\begin{tabular}[c]{@{}c@{}}Avg. \\ \#Unit Tests\end{tabular}   & 12.4   & 11.8  & 8.2   & -           \\
\midrule
\begin{tabular}[c]{@{}c@{}}Avg. \\ Coverage \end{tabular}       & 91.8   & 95.0  & 64.9$^{\ast}$ & -  \\
\bottomrule      
\end{tabular}
}
\caption{\devterm~Statistics. ${\dagger}$: \devterm~covers a range of domains including NLP, computer vision, deep learning, algorithm implementation, API applications, Database applications, web service (both frontend and backend), and general tools and utilities. ${\ddagger}$: We do not include unit testing for JavaScript as checking functional correctness is not applicable to pure static web pages. Correctness of page rendering and user interaction handling is checked using acceptance tests. ${\ast}$: Interfaces with thirty-party libraries that are not used to implement the designated functionalities are not supplied with test cases, resulting in relatively lower overall test coverage. 
}
\label{tab:repo stats}
\vspace{-5mm}
\end{table}

\subsection{Task 4: Acceptance Testing}

For this task, models are provided with the PRD, UML diagrams, and architecture design, with the option to include the implementation source code to generate acceptance test code.
Acceptance testing is critical to verify that the software adheres to requirements and operates effectively.
In the context of applications featuring command-line interfaces, acceptance tests interact with the software via shell commands, as specified in the PRD, and subsequently evaluate the accuracy of the output generated.
For libraries, acceptance tests are implemented through code that invokes the library's API, followed by assertions made on the responses of the API.
Applying this evaluative approach to LLMs provides valuable insights into their practical effectiveness and dependability within the domain of software development.

\textbf{Evaluation.} The evaluation of this task involves running the generated acceptance tests against the benchmark implementation code in the same testing framework developed for the evaluation for the implementation task.

In this phase, the Oracle Test methodology is employed to assess the accuracy of acceptance tests generated by the model, which are comprised of both test input and expected output.
The testing framework conducts an execution of the reference implementation of the software using the input devised by the model, subsequently comparing the software's actual output against the model-predicted output.
This approach facilitates an understanding of the extent to which LLMs can accurately interpret and predict the correct behavior of the subject software, as delineated in its design documentation.

\subsection{Task 5: Unit Testing}

In this phase, models are given the PRD, UML diagrams, and architecture design to generate unit test codes that facilitate a comprehensive understanding of the software. 
Unit testing serves as a fundamental approach to safeguard code integrity and operational accuracy.
Distinguished from broader acceptance testing, which focuses on the overall functionality and viability of the software, unit testing examines individual code segments for adherence to specified functionalities.
The increasing reliance on LLMs for streamlining software development processes underscores the criticality of their adeptness in writing effective unit tests, a competency integral to the software's reliability and overall robustness.

\textbf{Evaluation.} The evaluation of LLM-generated unit tests is conducted through their application on the reference source code, employing the previously delineated testing framework.
Similar to the acceptance testing evaluation, this phase leverages the Oracle Test, wherein the actual output of code units under specific test inputs is compared to anticipated outputs, as delineated by the oracle.

In addition, code coverage metrics are incorporated, providing a quantitative understanding of test comprehensiveness.
Utilizing statement coverage analysis tools integrated within the aforementioned testing frameworks, coverage is mathematically expressed as:
\begin{align*}
    \scalebox{0.9}{
    $\text{Coverage} = \left( \frac{\text{Number of Executed Statements}}{\text{Total Number of Statements}} \right) \times 100\%,$
    }
\end{align*}


where the number of executed statements denotes the count of distinct executable statements within the code that are executed at least once during the testing process, while the total number of statements represents the aggregate count of all executable statements present in the codebase that are subject to potential execution.

\subsection{Dataset}

The dataset construction process involved three phases: repository preparation, code cleanup, and document preparation. 
We first selected high-quality repositories from a GitHub dump, applying filters to ensure manageable complexity for evaluation.
Postgraduate student annotators then set up the environments, executed the code to verify functionality, and cleaned the repositories by removing unnecessary files and running or creating unit and acceptance tests to ensure oracle test standards. 
Finally, they prepared standard software design documents, including UML diagrams and architecture designs, following specific guidelines. 
The curated dataset consists of 22 repositories across Python, C/C++, Java, and JavaScript, with varying complexities and multi-file structures, as detailed in Table \ref{tab:repo stats}. 
Appendix \ref{appendix:dataset} provides more details.



\section{Experiments}


\subsection{Setup}

\paragraph{Models and the Baseline System}
We evaluate three prominent pre-trained model families with different model sizes, including both proprietary and open-source models: OpenAI GPT \citep{GPT}, CodeLlama \citep{roziere2023code}, DeepSeek-Coder \citep{guo2024deepseekcoder}.
Specifically, our experiments involve GPT-3.5-Turbo, GPT-4-Turbo from OpenAI GPT\footnote{We utilize \texttt{gpt-3.5-turbo-1106},\\ \texttt{gpt-4-0125-preview}, respectively.}, CodeLlama-Instruct 7B/13B/34B, and DeepSeek-Coder-Instruct models 1.3B/6.7B/33B.
Regarding the baseline system, we extend ChatDev \cite{qian2023communicative, qian2023experiential} for \devterm, adding support for UML diagrams, architecture design, environment setup, and multi-language execution, with structured PRDs and feedback to reduce hallucinations.
Further details are provided in the Appendix \ref{appendix:baseline}.

\begin{table}[htbp!]
    \centering
    \scalebox{0.75}{
    \begin{tabular}{l|cc}
        \toprule
        \textbf{Task}  & \multicolumn{2}{c}{\textbf{Implementation}} \\
        \midrule
        Evaluation Metric (\%)  & {\begin{tabular}[c]{@{}c@{}}Pass@\\ Accept. Test$^\P$\end{tabular}} & \multicolumn{1}{c}{\begin{tabular}[c]{@{}c@{}}Pass@ \\ Unit Test$^\P$\end{tabular}}      \\
        \midrule
        GPT-4-Turbo & & \\
        ~~~~\opt{No-Review} & 3.0 & 0.0\\
        ~~~~\opt{Normal-Review} & 3.0 & 0.0\\
        ~~~~\opt{Execution-Feedback} & 8.9 & 4.2\\
        \midrule
        CodeLlama-34B-Instruct & & \\
        ~~~~\opt{No-Review} & 0.0 & 1.4\\
        ~~~~\opt{Normal-Review}  & 0.0 & 1.4\\
        ~~~~\opt{Execution-Feedback} & 0.0 & 1.4 \\
        \midrule
        DeepSeek-Coder-33B-Instruct & & \\
        ~~~~\opt{No-Review} & 1.5 & 4.2 \\
        ~~~~\opt{Normal-Review} & 1.5 & 4.2 \\
        ~~~~\opt{Execution-Feedback} & 1.5 & 4.2 \\
        \bottomrule     
    \end{tabular}
    }
    \caption{Results of different prompting methods of the Implementation task on a subset of \devterm. $\P$: all results are averaged across all repositories and weighted by the number of code lines, which measures the difficulty of each repository. }
    \label{tab:impl_ablation}
\end{table}

\paragraph{Prompting Methods} 

In our baseline system, we explore three prompting methods: \textit{No-Review}, \textit{Normal-Review}, and \textit{Execution-Feedback}.
\textit{No-Review} represents a basic zero-shot prompting with built-in task prompts.
\textit{Normal-Review} involves a dual-role interaction, where the first role generates a solution and the second role reviews and, where necessary, corrects it. 
This mode is designed to evaluate the impact of review on model performance, in the absence of external inputs.
\textit{Execution-Feedback}, on the other hand, adds more dynamic interaction to the review process. 
This feedback includes runtime results, error messages, and performance metrics. 
Such information could enable the reviewing role to make more informed decisions, potentially leading to more accurate and effective solutions. For computational efficiency, we conduct a single review for all review-involved prompting methods.


\begin{table}[ht]
    \centering
    \scalebox{0.7}{ 
    \begin{tabular}{l|cc}
        \toprule
        \textbf{Task} & \multicolumn{2}{c}{\textbf{Unit Testing}} \\
        \midrule
        Evaluation Metric (\%)  & Oracle Test$^\S$ & Coverage$^\$$ \\
        \midrule
        GPT-4-Turbo & & \\
        ~~~~\opt{No-Review w/ Src Code} & 35.1 & 34.3 (54.8) \\
        ~~~~\opt{Normal-Review w/ Src Code} & 22.6 & 25.8 (68.7) \\
        \midrule
        CodeLlama-34B-Instruct & & \\
        ~~~~\opt{No-Review w/ Src Code} & 10.7 & 18.3 (73.2) \\
        ~~~~\opt{Normal-Review w/ Src Code} & 12.6 & 27.0 (72.1) \\
        \midrule
        DeepSeek-Coder-33B-Instruct & & \\
        ~~~~\opt{No-Review w/ Src Code} & 27.2 & 37.9 (75.8) \\
        ~~~~\opt{Normal-Review w/ Src Code} & 22.5 & 35.5 (71.0) \\
        \bottomrule     
    \end{tabular}
    }
    \caption{Results of different prompting methods of the Unit Testing task on a subset of \devterm. $\S$: the Oracle Test results are averaged across all repositories and weighted uniformly. $\$$: the results on the left side are averaged across all repositories and weighted uniformly, showing the overall scores. The results on the right side in the parenthesis are averaged across all \emph{valid} repositories and weighted uniformly, where models have generated \emph{executable} testing code.}
    \label{tab:unit ablation}
\end{table}

\begin{table}[ht]
    \centering
    \scalebox{0.8}{
    \begin{tabular}{l|c}
        \toprule
        \textbf{Task}  & \textbf{Accept. Testing} \\
        \midrule
        Evaluation Metric (\%) &  \multicolumn{1}{c}{Oracle Test$^\S$} \\
        \midrule
        GPT-4-Turbo & \\
        ~~~~\opt{No-Review} & 3.6 \\
        ~~~~\opt{Normal-Review} & 7.7 \\
        ~~~~\opt{Normal-Review w/ Src Code} & 14.9 \\
        \midrule
        CodeLlama-34B-Instruct & \\
        ~~~~\opt{No-Review} & 0.0 \\
        ~~~~\opt{Normal-Review} & 0.0 \\
        ~~~~\opt{Normal-Review w/ Src Code} & 0.0 \\
        \midrule
        DeepSeek-Coder-33B-Instruct & \\
        ~~~~\opt{No-Review} & 0.0 \\
        ~~~~\opt{Normal-Review} & 4.2 \\
        ~~~~\opt{Normal-Review w/ Src Code} & 15.6 \\
        \bottomrule     
    \end{tabular}
    }
    \caption{Results of different prompting methods of the Acceptance Testing task on a subset of \devterm. $\S$: all results are averaged across all repositories and weighted uniformly.}
    \label{tab:accept ablation}
\end{table}


\subsection{Main Results}

\paragraph{Results across prompting methods}
We first conduct experiments to examine the effects of different prompting methods on a subset of \devterm~using three representative models: GPT-4-Turbo, CodeLlama-34B-Instruct, and DeepSeek-Coder-33B-Instruct.


\begin{table*}[htbp!]
\centering
\scalebox{0.75}{
\begin{tabular}{l|c|cc|c|cr}
\toprule
\textbf{Task}                         & \textbf{Environment Setup}        & \multicolumn{2}{c|}{\textbf{Implementation}}                                                  & \textbf{Acceptance Testing} & \multicolumn{2}{c}{\textbf{Unit Testing}} \\
\midrule
Evaluation Metric (\%)  & {\begin{tabular}[c]{@{}c@{}}Pass@\\ Example Usage$^\S$\end{tabular}} & {\begin{tabular}[c]{@{}c@{}}Pass@\\ Accept. Test$^\P$\end{tabular}} & \multicolumn{1}{c|}{\begin{tabular}[c]{@{}c@{}}Pass@\\ Unit Test$^\P$\end{tabular}} & {Oracle Test}$^\S$       & {Oracle Test}$^\S$      & Coverage$^\$$      \\
\midrule
GPT-3.5-Turbo & \emph{33.3}  & 4.2 & 4.3 & 11.7 & 28.7 & 24.6 (61.4) \\
GPT-4-Turbo & \emph{\textbf{41.7}}  & \textbf{7.1} & \textbf{8.0} & \textbf{29.2} & \textbf{36.5} & 33.2  (66.3) \\
\midrule
CodeLlama-7B-Instruct & \emph{8.3}  & 0.0 & 0.0 & 0.0 & 3.0 & 3.6 (71.0) \\
CodeLlama-13B-Instruct & \emph{25.0}  & 0.6 & 0.0 & 0.0 & 5.1 & 8.6 (57.6) \\
CodeLlama-34B-Instruct & \emph{16.7}  & 0.6 & 0.5 & 4.5 & 21.1 & 25.4 (72.6) \\
\midrule
DeepSeek-Coder-1.3B-Instruct & \emph{8.3} & 0.0 & 0.1 & 0.0 & 5.6 & 2.7 (27.0)\\
DeepSeek-Coder-6.7B-Instruct & \emph{25.0} & 2.9 & 3.9 & 20.5$^\heartsuit$ & 23.5 & 28.2 (70.6) \\
DeepSeek-Coder-33B-Instruct & \emph{16.7} & 4.4 & 5.5 & 13.6 & 32.8 & 35.7 (79.4) \\
\bottomrule     
\end{tabular}
}
\caption{Tasks 2 to 5 results on \devterm.\emph{Italic figures}: test cases for the Environment Setup task are quite scarce compared to other tasks, therefore the results are more influenced by the randomness$^6$. $\S$: all results are averaged across all repositories and weighted uniformly. $\P$: all results are averaged across all repositories and weighted by the number of code lines. $\$$: the results on the left side are averaged across all repositories and weighted uniformly, showing the overall scores. The results on the right side in the parenthesis are averaged across all \emph{valid} repositories and weighted uniformly, where models have generated \emph{executable} testing code. $\heartsuit$: the model has generated meaningless but executable testing code. }
\label{tab:main results}
\end{table*}

Table \ref{tab:impl_ablation} illustrates the results of the implementation task for our study.
In general, \opt{Execution-Feedback} leads to the optimal performance, where GPT-4-Turbo benefits the most, especially on acceptance tests.
In contrast, CodeLlama-34B-Instruct and DeepSeek-Coder-33B exhibited no improvement with any review process.
We note that the efficacy of the review process could be understated as our automated testing is too rigorous and sparse to reflect the improvements.
We observe that, despite no substantial improvements on reference tests, the code quality notably improved with \opt{Execution-Feedback} prompt.
However, there is no significant improvements using the \opt{Normal-Review} setting compared with the \opt{No-Review} setting.
Models consistently provide unhelpful suggestions, such as the addition of unnecessary error handling or reorganization.
This indicates that the models are unable to comprehend complicated code by merely reading it, lacking external knowledge like execution feedback.

For the testing tasks, which are relatively easier than the implementation, there are various observations.
In Table \ref{tab:unit ablation}, we find no clear evidence that the review process brings stable benefits to unit testing.
However, on the acceptance testing in Table \ref{tab:accept ablation}, \opt{Normal-Review} brings enhancement to the performance.
The degradation in unit testing performance with review mainly stems from extended input length challenging models' long-context comprehension. Imprecise reviewer suggestions may also reduce output quality.
Regarding the visibility of implementation source code, it is common practice not to expose source code and execution feedback for acceptance testing, while it's allowed to employ the source code as input for unit testing.
As shown in Table \ref{tab:accept ablation},  models can barely generate executable acceptance testing code and incorporating source code as additional input dramatically increases the performance.



\paragraph{Results across models}

Table \ref{tab:main results} illustrates the main results on \devterm~with optimal prompting methods applied for each task.\footnote{
\opt{Execution-Feedback} is used for Environment Setup and Implementation; \opt{Normal-Review w/ Src Code} for Acceptance Testing; \opt{No-Review w/ Src Code} for Unit Testing as \opt{Normal-Review w/ Src Code} shows no clear advantage.
}
We find that GPT-4-Turbo demonstrates superior performance compared to other models, while all models are far from satisfactory.
\devterm~can effectively distinguish between models of varying capabilities.
Smaller models, such as CodeLlama-7B/13B-Instruct and DeepSeek-Coder-1.3B-Instruct, demonstrate inherent limitations, frequently unable to generate syntactically accurate code or follow the instructions.
These models tend to generate mere code skeletons or fill the function body with only comments.
Larger open-sourced models and GPT models, while generating more reasonable code, still struggle with the subtleties of complex code structure and logic, such as variable type conversion, function arguments and object-oriented classes.


\paragraph{Results across tasks}

Generally, the models' performances on the implementation task are all below the 10\% pass rate. 
The highest-performing model, GPT-4-Turbo, registers only a 7.1\% pass rate on reference acceptance tests and 8.0\% on unit tests, while some other models score zero, failing all reference tests.
Despite prior research such as HumanEval~\citep{chen2021evaluating} indicating that models can manage simple code-writing tasks, substantial challenges remain in more complex coding scenarios within \devterm.
We break down the implementation results into different languages in Figure \ref{fig:language_performance} and find that models particularly struggle in handling Java and C/C++, whose stringent syntax requirements tend to magnify the models' deficiencies in managing intricate details.
This points to the necessity for more enriched and diverse training data across programming languages to bridge this competency gap.

Compared with the implementation task, other tasks are relatively simple but still challenging.
Regarding the environment setup task, we note that the test cases for the environment setup task are quite scarce compared to other tasks\footnote{
Except for C/C++ and easy Python repositories that are absent, only 12 repositories are involved in the environment setup task.
We will resolve this issue in future work.
However, \devterm~contains rich test cases for other tasks.
Table \ref{tab:repo stats} evidences that \devterm~features fruitful tests for the implementation task. 
For the testing generation tasks, our prompts depict fine-grained requirements and ensure the quantity of generated testing cases.
}, therefore the results are more influenced by the randomness.  
Roughly speaking, GPT-4-Turbo reaches a 41.7\% pass rate in building environments, while open-sourced models largely fall behind it.
For the testing tasks, GPT-4-Turbo still obtains the highest scores and open-sourced models perform worse.
We identify an outlier that DeepSeek-Coder-6.7B-Instruct achieves the highest score among open-sourced models and approach GPT-4-Turbo in acceptance testing.
The model somehow \emph{cheats} on this task by  generating meaningless but executable testing code (detail in Section \ref{sec:Limitation in test}).
With respect to the unit testing, the overall Oracle Test and Coverage scores are quite low, which are averaged across all repositories. 
However, for those generated testing codes that can be successfully executed (obtain non-zero Oracle Test score), the coverage scores are relatively high, suggesting the models' promising capability on this problem.

\paragraph{Results on software design}
Table \ref{tab:design results} shows the results of software design using GPT-4-Turbo as the Judge and GPT-3.5-Turbo as the baseline for comparison.
More details on how we evaluate the software design are found in Appendix \ref{appedix:eval of design}.
GPT-4-Turbo dominantly outperforms GPT-3.5-Turbo with extraordinarily high win rates in all cases.
Regarding the open-sourced models, as the size increases, models consistently produce higher quality design documents on both metrics, while they are relatively inferior on \emph{faithfulness}.

We compare the LLM-as-a-Judge results with human majority annotations.
Low agreements are observed with tie considered, which aligns with previous studies \citep{zheng2023judging}.
It is reasonable as a tie is hard to define and judge, especially for highly complicated and structured software design documents.
Without tie, GPT-4-Turbo reaches 79.2\% and 83.2\% agreements on the \emph{general principles} and \emph{faithfulness} metrics, respectively.
This means GPT-4-Turbo's judgments align with the majority of humans and could serve as a good alternative for automated software design evaluation.

\begin{table}[htbp!]
    \centering
    \scalebox{0.75}{
    \begin{tabular}{l|rr|rr}
        \toprule
         & \multicolumn{2}{c|}{\textbf{w/ Tie}}                          & \multicolumn{2}{c}{\textbf{w/o Tie}}   \\ 
                    & \multicolumn{1}{c}{G$^\dag$} & \multicolumn{1}{c|}{F$^\ddag$} & \multicolumn{1}{c}{G} & \multicolumn{1}{c}{F} \\
        \midrule
        GPT-4-Turbo & \textbf{97.9} & \textbf{97.9} & \textbf{100.0} & \textbf{100.0}      \\
        \midrule
        CodeLlama-7B-Instruct & 4.2 & 8.3 & 4.2 & 4.5  \\
        CodeLlama-13B-Instruct & 18.8 & 14.6 & 10.5 & 5.3  \\
        CodeLlama-34B-Instruct & 39.6 & 33.3 & 33.3 & 21.4   \\
        \midrule
        DeepSeek-Coder-1.3B-Instruct & 16.7 & 16.7 &  5.5  & 5.6  \\
        DeepSeek-Coder-6.7B-Instruct & 35.4 & 35.4 &  31.6  & 29.4  \\
        DeepSeek-Coder-33B-Instruct  & 52.1 & 50.0 &  53.8  & 50.0 \\
        \midrule
        \midrule
        Agree w/ Human Majority  &  60.4   &  51.6  & 79.2 & 83.2 \\
        \bottomrule
    \end{tabular}
    }
    \caption{Win rate of pairwise comparison against GPT-3.5-Turbo on Software Desgin on a subset of \devterm~. Results are averaged across different repositories and sub-tasks uniformly. $\dag$: \emph{general principles}. $\ddag$: \emph{faithfulness}. w/ Tie: inconsistent results are considered as a tie. We also report agreement with Human Majority.}
    \label{tab:design results}
\end{table}

\subsection{Analysis}
Our experiments reveal several challenges faced by LLMs, such as difficulties in generating accurate Makefiles and Gradle files, handling function redefinitions, and managing file references in multi-file repositories. 
Models also struggle with correct function parameter usage, naming conventions, type handling, and managing variable scope. Furthermore, they frequently fabricate variables or misinterpret data files, leading to hallucination issues.
For more detailed experimental findings and analysis, please refer to the Appendix \ref{Appendix: Discussions}.

\section{Conclusion}
The \devterm~framework presents a leap forward in studying LLMs within the domain of automated software development.
By employing a multi-stage evaluation process, \devterm~comprehensively assesses LLMs across a spectrum of tasks including design, environment setup, implementation, and testing.
Empirical findings reveal that pre-trained models like GPT-4-Turbo are still confronted with substantial challenges within \devterm.
Through analysis, we identify models' limitations in understanding the complex repository structures and handling the nuanced demands of comprehensive software development.
These insights elucidate critical pathways for future model development.

\section{Limitations}

One limitation of our work is the limited number of repositories used in the study, with only 22 curated examples across four programming languages. 
This relatively small dataset may not fully capture the wide variety of challenges and complexities present in real-world software development. 
Although we selected diverse repositories to represent different domains and programming paradigms, a broader and more extensive collection of repositories would provide a more comprehensive evaluation of LLM performance. 
Future work could address this by incorporating additional repositories to better generalize the findings.

\bibliography{reference}

\clearpage
\appendix
\newpage
\startcontents[appendices]
\section*{Appendix Table of Contents}
\printcontents[appendices]{}{0}{\large}
\newpage

\section*{Appendix}
\section{Software Engineering Tasks with a Running Example}
\label{appendix:se-tasks}

We describe concepts of software engineering tasks using one of the subjects of \devterm, named Actor Relationship Game.
The Actor Relationship Game is a Java-based application that allows users to explore connections between popular actors through their movie collaborations, using data from The Movie Database (TMDB) API. It constructs an actor graph and identifies the shortest path of relationships between any two actors.

\begin{figure*}[ht!]
    \centering
    \includegraphics[width=0.5\linewidth]{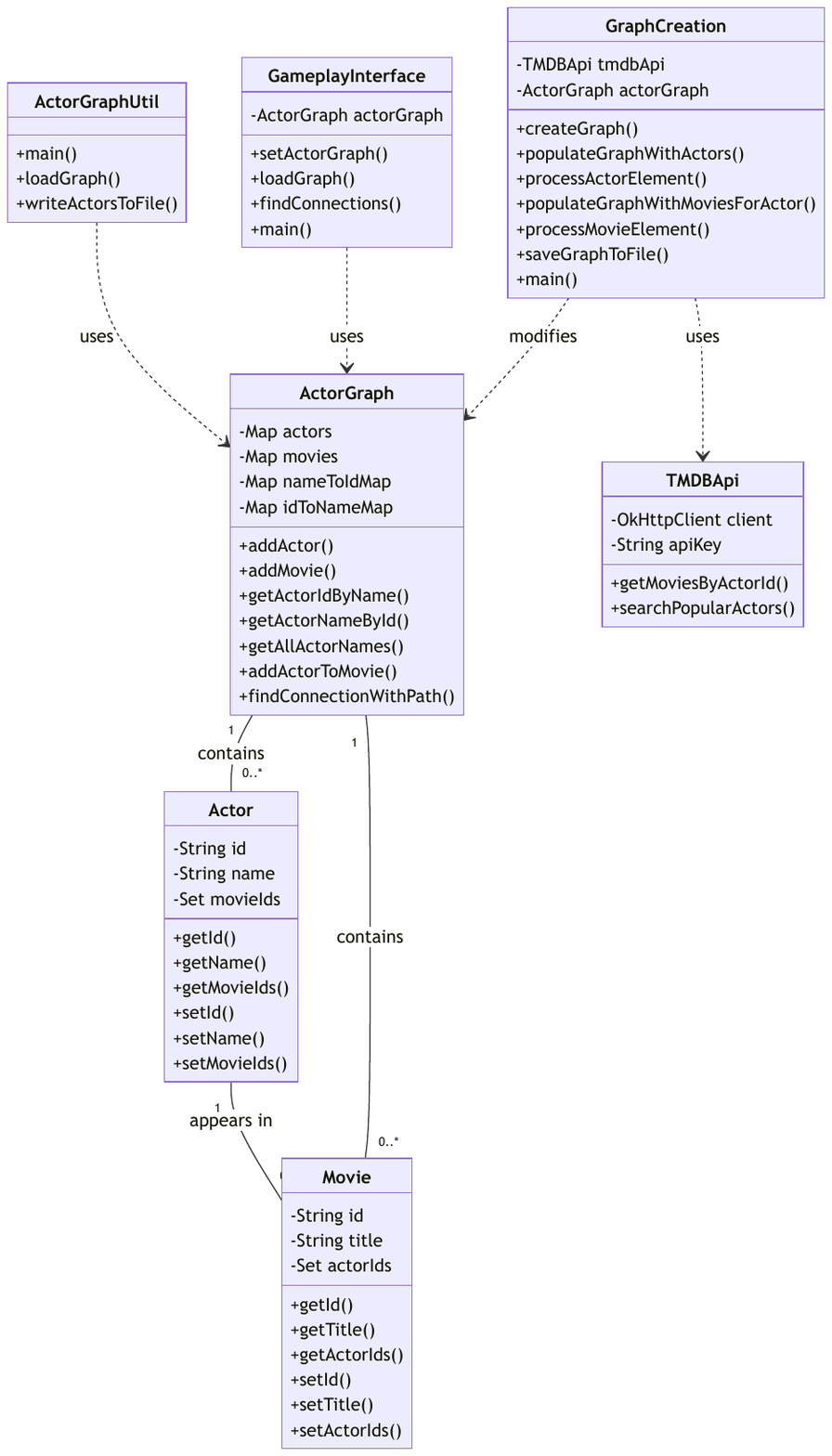}
    \caption{UML Class Diagram for the Example Repository}
    \label{fig:uml-class}
\end{figure*}

\begin{figure*}[ht!]
    \centering
    \includegraphics[width=0.7\linewidth]{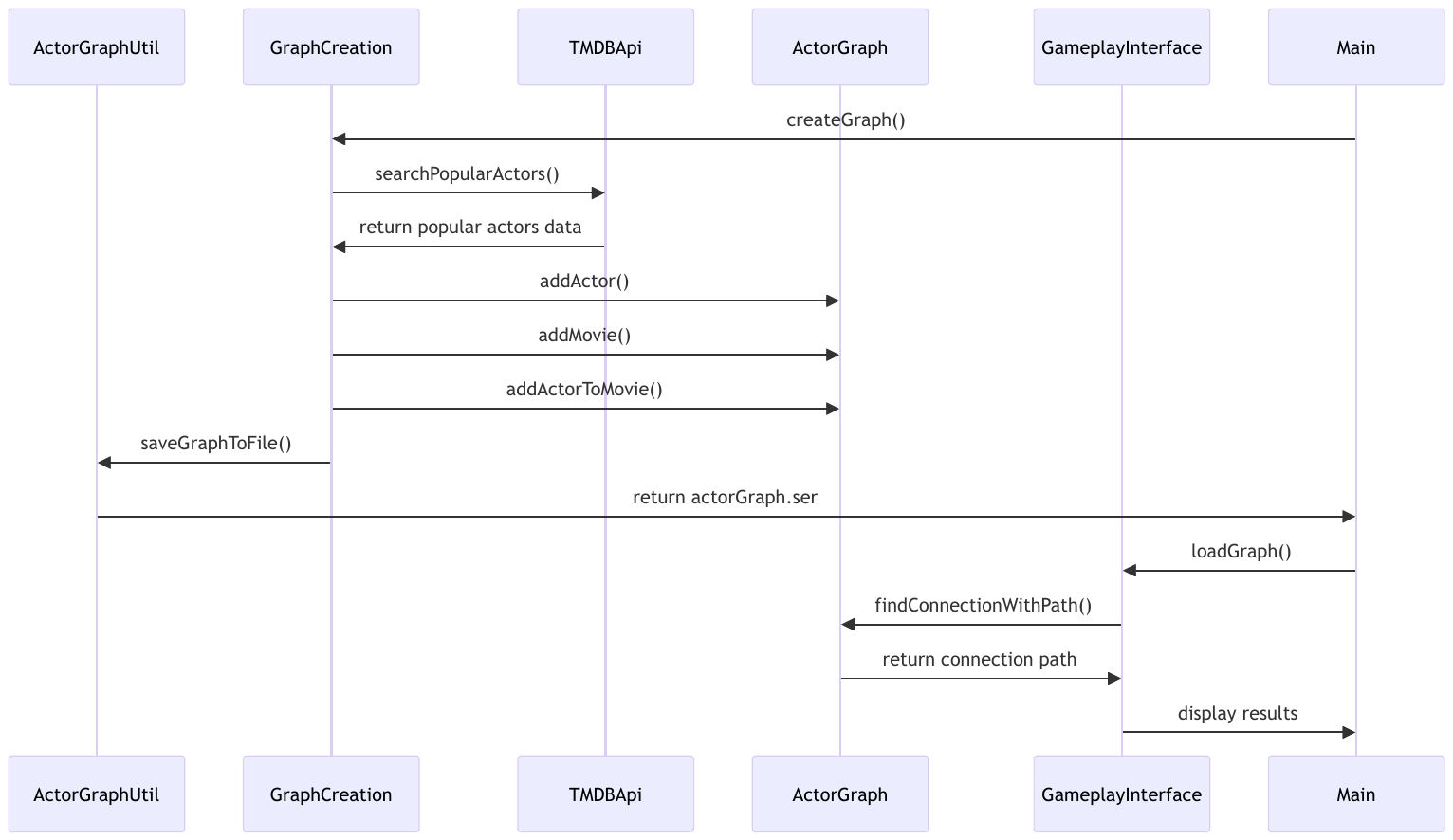}
    \caption{UML Sequence Diagram for the Example Repository}
    \label{fig:uml-sequence}
\end{figure*}

\subsection{Software Design}
\label{appendix:software-design}

Software design is the process by which an agent creates a specification of a software artifact, intended to accomplish goals, using a set of primitive components and subject to constraints.
It is a phase in the software development lifecycle that bridges the gap between software requirements analysis and the actual implementation of the software system.

During the software design phase, software engineers or designers define the way a software application will work to meet the specified requirements in the form of a Product Requirement Document (PRD).
They create diagrams that determine the data structures, software architecture, interface designs, and module specifications with Unified Markup Language (UML) diagrams.

A good software design is crucial as it impacts the quality, maintainability, performance, scalability, and robustness of the software product.
It facilitates a smoother implementation phase, allows for better understanding and communication among team members, and helps in identifying potential issues early in the development process.

\subsubsection{Class Diagrams}
\label{appendix:class-diagrams}

Class diagrams are a cornerstone of object-oriented design, offering a static snapshot of the system structure. These diagrams illustrate the classes within the system, their attributes, methods, and the relationships among the classes, such as inheritance and associations. Class diagrams are instrumental in providing an abstract representation of the system’s components and their interactions, facilitating a deeper understanding of the software’s overall architecture and design patterns.

Figure~\ref{fig:uml-class} shows the class diagram of the Actor Relationship Game repository.
This UML class diagram delineates the architecture of a system designed to model and analyze the network of relationships between actors and movies through an \texttt{Actor Graph}.
It encompasses classes such as `Actor` and `Movie` to represent individual entities, alongside an \texttt{ActorGraph} class that serves as a repository and management layer for these entities and their associations.
Utility and operational classes like \texttt{ActorGraphUtil}, \texttt{GameplayInterface}, and \texttt{GraphCreation} provide mechanisms for manipulating the graph—ranging from data ingestion using the \texttt{TMDBApi} to utility functions and gameplay interfaces that leverage the graph for various applications.
The relationships between classes, including associations, aggregations, and dependencies, are meticulously outlined to depict interactions such as actors appearing in movies and the construction and utilization of the actor-movie graph for finding connections and supporting gameplay or analysis tasks.

\subsubsection{Sequence Diagrams}
\label{appendix:sequence-diagrams}

Sequence diagrams, different than class diagrams, focus on the dynamic aspects of the system. They depict how objects interact with each other across time, outlining the sequence of messages exchanged between objects to accomplish a specific functionality or process within the system. Sequence diagrams are invaluable for visualizing and analyzing the flow of operations, timing constraints, and the interaction patterns among system components, making them essential for detailed behavioral analysis.

The sequence diagram of \texttt{Actor Relationship Game} repository is illustrated in Figure~\ref{fig:uml-sequence}.
This diagram illustrates the flow of operations for creating, populating, and utilizing an actor-movie graph.
Initially, the \texttt{Main} function triggers the graph creation process by calling \texttt{createGraph()} on the \texttt{GraphCreation} module, which then interacts with the \texttt{TMDBApi} to fetch popular actors' data.
Upon receiving this data, \texttt{GraphCreation} populates the \texttt{ActorGraph} with actors, movies, and their associations.
After constructing the graph, \texttt{GraphCreation} delegates the responsibility of saving this graph to a file to \texttt{ActorGraphUtil}, which then returns a serialized file (\texttt{actorGraph.ser}) back to \texttt{Main}.
Subsequently, \texttt{Main} instructs the \texttt{GameplayInterface} to load this graph and use it to find connections between actors via \texttt{findConnectionWithPath()}, a method in \texttt{ActorGraph}.
The path found is then returned to \texttt{GameplayInterface}, which finally displays the results back in the \texttt{Main} function.
This sequence encapsulates a complete lifecycle from graph creation, through data population and serialization, to utilization for finding actor connections, showcasing a systematic approach to managing and analyzing actor-movie relationships.

\subsubsection{Architecture Design}
\label{appendix:arch-design}

Architecture design using file tree representation refers to a method of visualizing and organizing the structural layout of a software system's components in a hierarchical format. This approach delineates the organization of software modules, packages, libraries, and other assets in a tree-like structure, where each node represents a file or a directory containing more files or directories. Such a representation is crucial in conveying the architectural blueprint of a software project, illustrating how its various parts are interrelated.

The text-based representation of the file tree for the Actor Relationship Game repository, including test classes for each Java class, is shown as below.

{\scriptsize\dirtree{%
.1 /.
.2 .gitignore.
.2 src.
.3 acceptanceTest.
.4 java.
.5 Actor\_relationship\_game.
.6 ARG\_AcceptanceTest.java.
.3 main.
.4 java.
.5 Actor\_relationship\_game.
.6 Actor.java.
.6 Movie.java.
.6 ActorGraph.java.
.6 ActorGraphUtil.java.
.6 GameplayInterface.java.
.6 GraphCreation.java.
.6 TMDBApi.java.
.3 test.
.4 java.
.5 Actor\_relationship\_game.
.6 ActorTest.java.
.6 MovieTest.java.
.6 ActorGraphTest.java.
.6 ActorGraphUtilTest.java.
.6 GameplayInterfaceTest.java.
.6 GraphCreationTest.java.
.6 TMDBApiTest.java.
.2 build.gradle.
.2 README.md.
}}

\subsection{Software Development}
\label{appendix:software-development}

Software development is the comprehensive process of programming, documenting, optimization, and fixing involved in creating and maintaining applications, frameworks, or other software components.
It encompasses all the activities that result in software products and involves a series of steps known as the software development lifecycle (SDLC).

\begin{itemize}
    \item \textbf{Environment Setup} is the process of preparing and configuring the necessary hardware and software tools required to build and run software applications. This setup is crucial to provide a consistent, controlled, and efficient workspace for developers to code, test, and deploy their applications. The environment can be set up on an individual's local machine, on a remote server, or in a containerized environment.
    \item \textbf{Implementation} is when developers write code according to the software design documents, using programming languages and tools suitable for the repository.
 \end{itemize}

\subsection{Quality Assurance}
\label{appendix:quality-assurance}

Quality Assurance (QA) is the systematic process of ensuring that the software being developed meets the specified quality standards and requirements before it is released.

Software testing is an integral part of QA; involves the execution of a software component or system component to evaluate one or more properties of interest. Software testing typically includes:

\begin{itemize}
    \item \textbf{Unit Testing} is the process of testing individual units or components of a software application to ensure their behaviors. A unit is the smallest testable part of any software and usually has one or a few inputs and usually a single output. In procedural programming, a unit could be an entire module, but it is more commonly an individual function or procedure.
    \item \textbf{Acceptance Testing} is a level of software testing where a system is tested for acceptability. It provides the final assurance that the software meets the PRD and is ready for use by end-users.
\end{itemize}

Code Listing~\ref{lst:ut} is part of the unit test suite for the Actor Relationship Game Repository, specifically designed to validate the functionality of the \texttt{Actor} class.
Using the JUnit framework, it defines two test cases: \texttt{testActorIdAndName} and \texttt{testMovieIds}.
The first test, \texttt{testActorIdAndName}, instantiates an  \texttt{Actor} object with a specific ID and name ("101" and "John Doe", respectively) and asserts that the \texttt{getId()} and \texttt{getName()} methods correctly return these values, ensuring the actor's identity is accurately stored and retrievable.
The second test, \texttt{testMovieIds}, creates another \texttt{Actor} object and adds two movie IDs ("201" and "202") to the actor's list of movie IDs.
It then verifies that these movie IDs are indeed associated with the actor by checking if the actor's \texttt{getMovieIds()} set contains the added IDs. 
Together, these tests check the integrity of the \texttt{Actor} class's basic functionalities: maintaining an actor's identity and managing their associated movie IDs.

\begin{lstlisting}[language = Java , frame = tb ,escapeinside={(*@}{@*)}, caption={Example Unit Test}, label=lst:ut]
package Actor_relationship_game;

import org.junit.jupiter.api.Test;
import static org.junit.jupiter.api.Assertions.*;

class ActorTest {
    @Test
    void testActorIdAndName() {
        Actor actor = new Actor("101", "John Doe");
        assertEquals("101", actor.getId());
        assertEquals("John Doe", actor.getName());
    }

    @Test
    void testMovieIds() {
        Actor actor = new Actor("102", "Jane Smith");
        actor.getMovieIds().add("201");
        actor.getMovieIds().add("202");
        assertTrue(actor.getMovieIds().contains("201"));
        assertTrue(actor.getMovieIds().contains("202"));
    }
}
\end{lstlisting}

In Code Listing~\ref{lst:at}, the example acceptance test is designed to verify the functionality of generating and comparing actor lists from graph data.
It employs the \texttt{runGradleTask} method to execute specific Gradle tasks for creating graph data and generating actor lists into specified file paths, using parameters for file names to differentiate between reference and test data.
The test first runs the \texttt{runGraphCreation} task with paths for both reference and test graphs, followed by the \texttt{runActorGraphUtil} task to generate actor lists from these graphs into specified file paths.
Once the actor lists are generated, the test reads lines from both the reference and test actor list files and then iterates through each line in the reference actor list, followed by an assertion that each actor from the reference list is also present in the test list.
This process effectively checks the integrity and consistency of the actor list generation feature by ensuring that the test actor list replicates the reference list accurately, thereby validating the application's capability to process and output graph-related data correctly.

\begin{lstlisting}[language = Java , frame = tb ,escapeinside={(*@}{@*)}, caption={Example Acceptance Test}, label=lst:at]
@Test
public void testActorList() throws IOException,InterruptedException{
    runGradleTask("runGraphCreation -PfileName="+referenceGraphPath);
    runGradleTask("runGraphCreation -PfileName="+testGraphPath);
    runGradleTask("runActorGraphUtil -PgraphPath="+referenceGraphPath+" -PfilePath="+referenceActorPath);
    runGradleTask("runActorGraphUtil -PgraphPath="+testGraphPath+" -PfilePath="+testActorPath);


    List<String> referenceLines = Files.readAllLines(Paths.get(referenceActorPath));
    List<String> testLines = Files.readAllLines(Paths.get(testActorPath));

    for (String referenceLine:referenceLines){
        assertTrue(containsLine(testLines,referenceLine));
    }
}
\end{lstlisting}

\section{Dataset}
\label{appendix:dataset}
\subsection{Dataset Construction}

The data preparation process in our work consists of three distinct phases: repository preparation, code cleanup, and document preparation.

The initial phase, repository preparation, involves selecting high-quality, well-structured candidate repositories from a GitHub dump. 
Recognizing the impracticality of constructing a repository from scratch, we employed a filtering process to identify suitable candidates. 
Moreover, we imposed a constraint on the total number of lines of code to ensure the repositories' complexity remains manageable, facilitating the evaluation of current LLMs.

During the code cleanup phase, our postgraduate student annotators were tasked with setting up the required environment as stipulated in the repositories' README files.
 They then executed the code to verify its functionality.
 Following this sanity check, the annotators were instructed to meticulously refine the code repositories. 
 This refinement included the removal of unnecessary auxiliary files. 
 To ascertain code quality, the annotators were also required to run existing unit and acceptance tests, or to develop additional tests, ensuring they meet the standards of the oracle test and achieve satisfactory coverage.

The final phase, document preparation, involved the creation of standard software design documents, namely, UML class and sequence diagrams, and architecture designs, for each repository. 
We provide annotators with specific guidelines and templates for these documents. 
The annotators were responsible for ensuring that these design documents corresponds accurately and cohesively with the respective code repositories.

\subsection{Dataset Statistics}

In Table \ref{tab:repo stats}, we present an exhaustive statistical breakdown of our datasets.
\devterm~contains a collection of 22 curated repositories, spanning across four widely-used programming languages (Python, C/C++, Java, JavaScript) and a diverse range of domains.
The dataset is characterized by its multi-file structure.
The Python repositories in our dataset are relatively straightforward, with each repository comprising approximately two files and an average of 276 lines of code.
In contrast, the repositories pertaining to statically-typed programming languages, namely C/C++ and Java, are more complex, featuring an increased count of code files and lines.
For JavaScript, our usage of the Vue.js framework necessitates that models adeptly navigate the framework's templates and development paradigms.
Consequently, JavaScript repositories exhibit the highest number of code files and lines, posing a substantial challenge for LLMs.
Additionally, we have prepared extensive reference acceptance and unit tests for each repository to facilitate rigorous evaluation of the implementation task.

\section{The Baseline System}
\label{appendix:baseline}

We introduce our baseline system formulated for \devterm, building upon the foundations of ChatDev \cite{qian2023communicative, qian2023experiential}. 
ChatDev is a virtual, chat-powered software development system that adheres to the conventional waterfall model. 
It bifurcates the development process into four primary tasks (dubbed as phrases in ChatDev): design, coding, testing, and documentation. 
Within this system, multiple LLM agents assume diverse roles such as programmers, reviewers, and testers, pertinent to each phase. 
ChatDev is characterized by its utilization of a chat chain mechanism, which segments each phase into smaller, atomic tasks. 
This approach enables context-sensitive, multi-turn dialogues between two distinct roles, facilitating the proposal and validation of solutions for individual tasks.

In contrast to ChatDev, our development of baseline system incorporates several features and enhancements. 
We have restructured the task design to align closely with the evaluation criteria of \devterm. 
Specifically, this includes the integration of comprehensive input to the system, exemplified by well-structured PRDs. 
This integration is crucial in addressing and examining the issue of hallucination in controlled experimental settings. 
Moreover, our baseline system expands upon the capabilities of ChatDev, supporting a wider range of tasks, including standard Object-Oriented programming designs (UML class and sequence diagrams), repository planning (architecture design), environment setup, and acceptance testing.
A significant advancement in our baseline system is its compatibility with multiple programming languages and their corresponding runtime environments. 
This feature is coupled with the provision of comprehensive execution feedback to the system. 

\paragraph{Implementation Details}
We utilize LMDeploy for the deployment of CodeLlama and DeepSeek-Coder models.\footnote{\url{https://github.com/InternLM/lmdeploy}}
Acknowledging the potential for extensive input context in \devterm~tasks, we configure the context length to 32K for these models.
For the Software Design task, we set the temperature parameter to 0.2, while for the remaining four tasks, we use a temperature of 0.
Other hyperparameters in the experiment are maintained at default settings.
All code-related tasks are rigorously evaluated in an isolated sandbox environment, utilizing Docker technology.

\section{Software Design Evaluation}
\label{appedix:eval of design}

\begin{figure*}[ht!]
    \centering
    \includegraphics[width=0.9\linewidth]{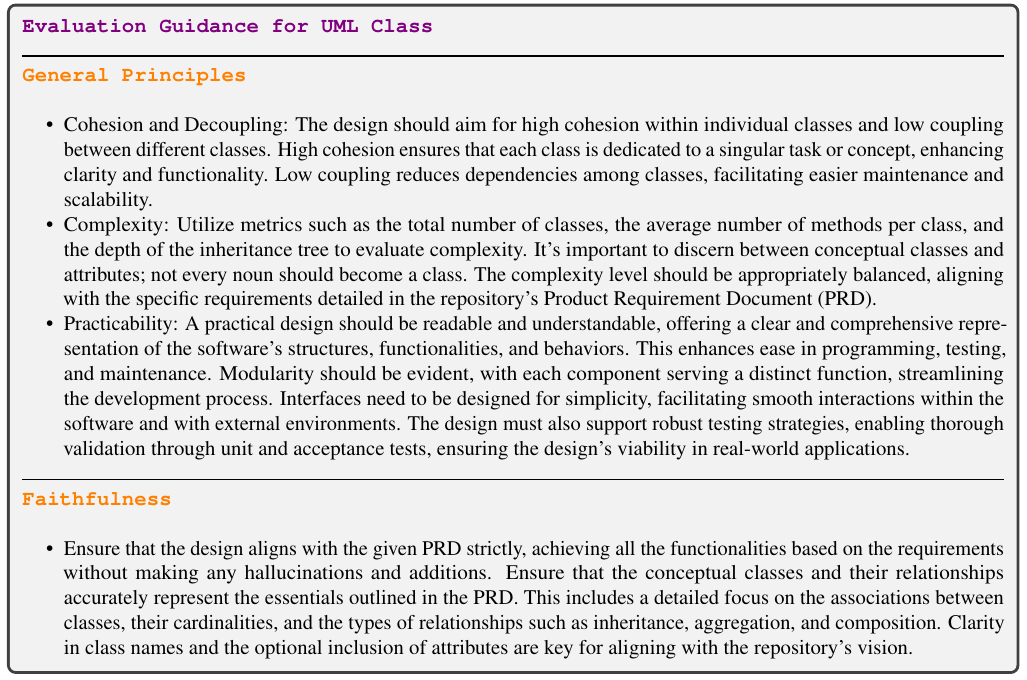}
    \caption{Evaluation Guidance for UML Class.}
    \label{fig:uml_class_metric}
\end{figure*}

\begin{figure*}[ht!]
    \centering
    \includegraphics[width=0.9\linewidth]{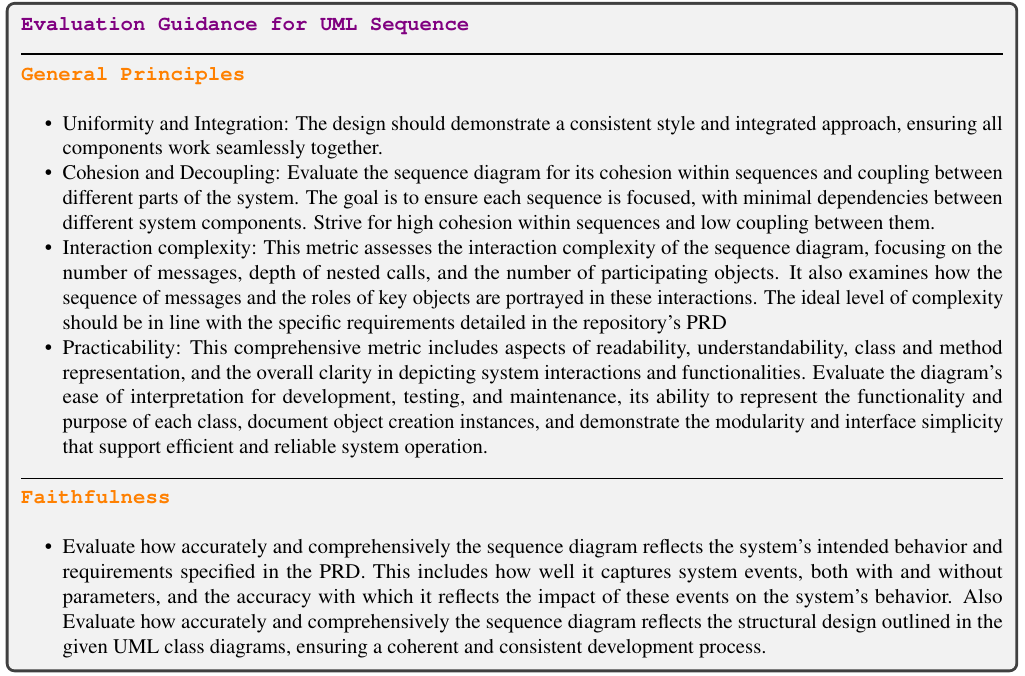}
    \caption{Evaluation Guidance for UML Sequence.}
    \label{fig:uml_seq_metric}
\end{figure*}

\begin{figure*}[ht!]
    \centering
    \includegraphics[width=0.9\linewidth]{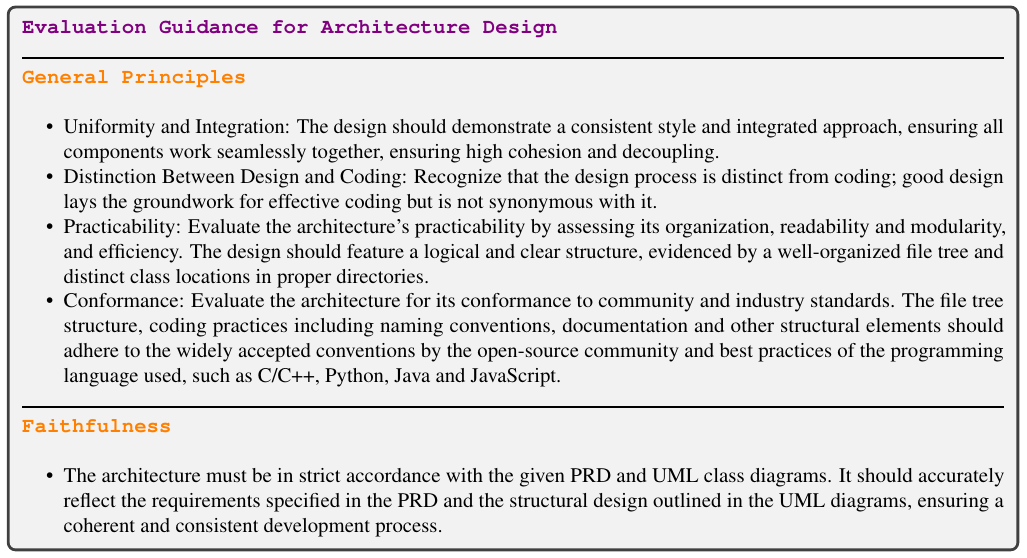}
    \caption{Evaluation Guidance for Architecture Design.}
    \label{fig:archi_metric}
\end{figure*}

We follow previous work \cite{zheng2023judging} to conduct a pairwise comparison to determine which response is better, focusing on the metrics of \emph{general principles} and \emph{faithfulness} (see the corresponding prompts in Figures~\ref{fig:uml_class_metric},~\ref{fig:uml_seq_metric},~\ref{fig:archi_metric}).
To reduce the expenditure of the OpenAI GPT API and human effort, the scope of our evaluation was confined to a subset of our dataset.
This process involves 192 pairs across eight repositories, eight models, and three sub-tasks.
Regarding the LLM judge, we use GPT-4-Turbo as the judge and GPT-3.5-Turbo as the baseline model.
To mitigate the issue of position bias \citep{zheng2023judging,shi2024judgingjudgessystematicinvestigation}, i.e., LLM judges preferring response at a certain position regardless of the content, the evaluation was executed in a \emph{dual} mode, evaluating each pair twice in different orders (384 pairs in total), with inconsistent decisions being considered as a tie.
For the human evaluation, we shuffle the order of two responses and annotate each pair thrice to obtain the \emph{human majority}.

\begin{figure*}[ht!]
    \centering
    \includegraphics[width=0.9\linewidth]{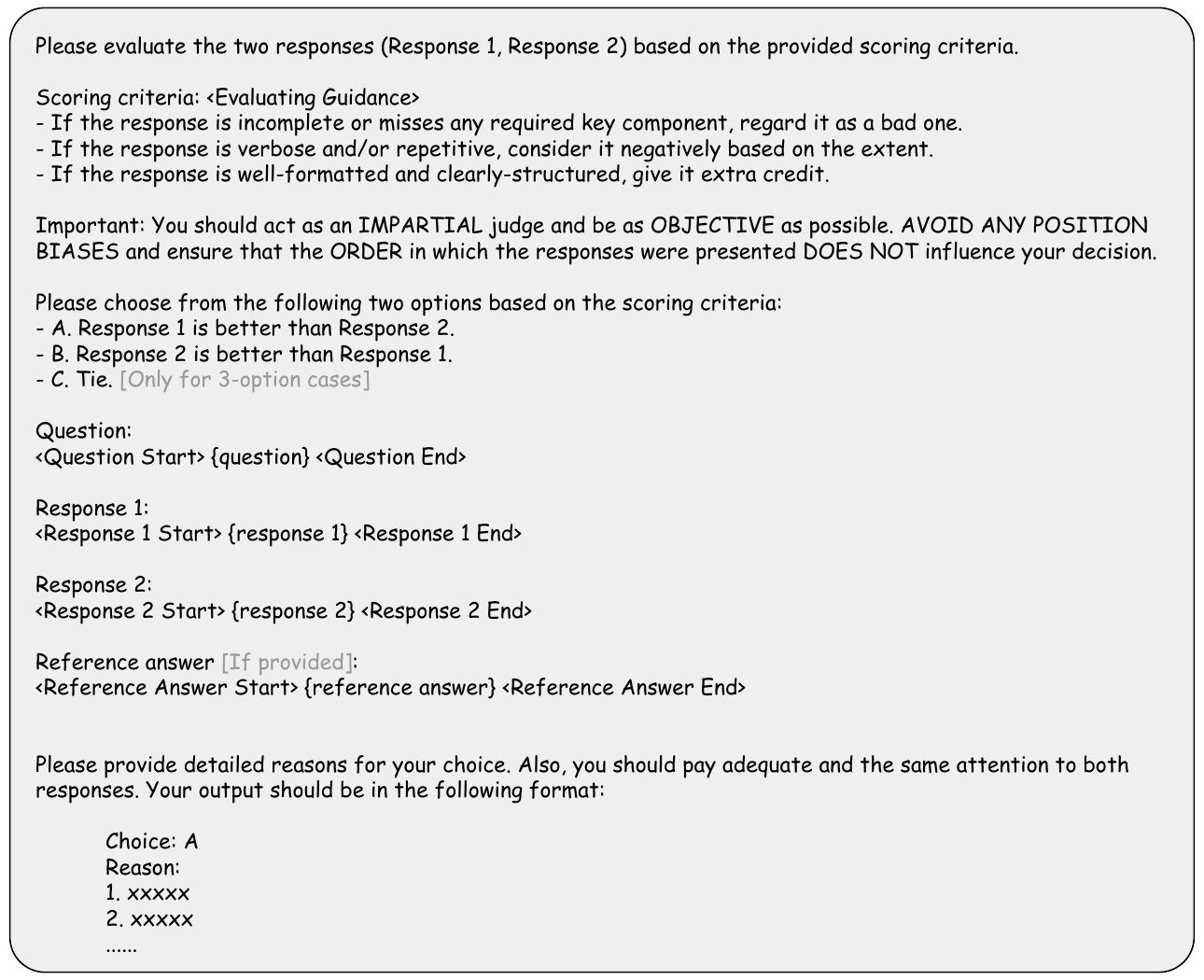}
    \caption{LLM-as-a-Judge prompt for software design evaluation}
    \label{fig:judge_prompt}
\end{figure*}

The customized LLM-as-a-Judge prompt for evaluating software design is detailed in Fig. \ref{fig:judge_prompt}, structured to facilitate pairwise comparisons in accordance with predefined evaluation guidelines. The LLM's judgments are extracted by employing regular expressions to identify the selection specified after "Choice:" within the judges' responses.

\begin{figure*}
    \centering
    \includegraphics[width=0.8\linewidth]{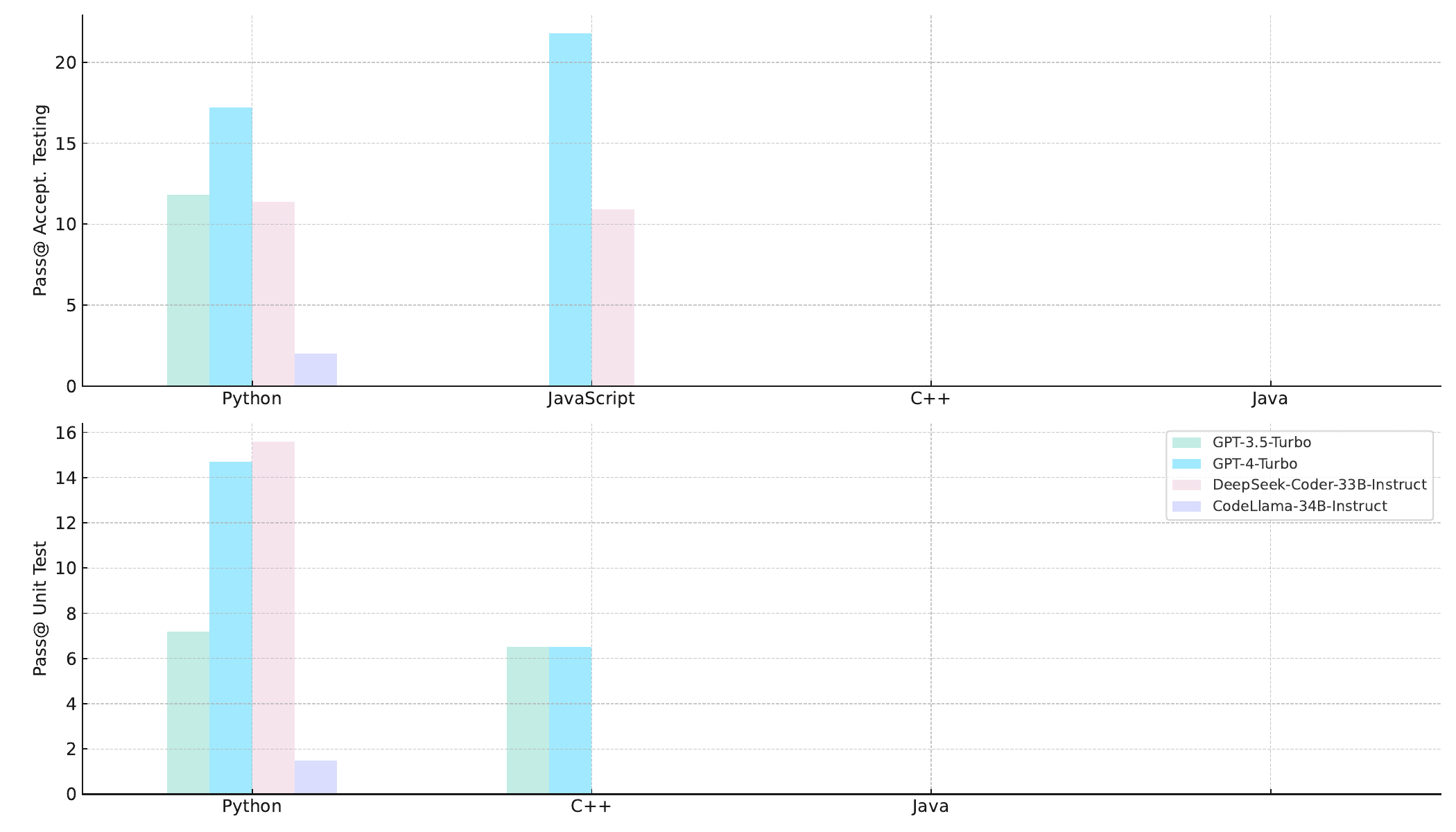}
    \caption{Performance Break Down on Different Languages. The results are averaged across all repositories and weighted by the number of code lines}
    \label{fig:language_performance}
\end{figure*}

\section{Experimental Discussions}
\label{Appendix: Discussions}

\subsection{Model Capacity}
\paragraph{Challenges in Creating C/C++ Makefile and Java Gradle}
LLMs often face challenges in generating accurate Makefile for C/C++ and Gradle files for Java.
Frequently, the generated files are deficient in critical components like source code files, necessary dependencies and essential tasks. 
In C/C++ and Java repositories, approximately 90\% of compilation and execution errors can be attributed to these issues. 
We find that even GPT-4-Turbo occasionally fail in basic syntax errors on compilation files.
This is potentially caused by insufficient training data related to these compilation tools.


\paragraph{Function Redefinition in Multi-file Repositories}
Models face significant challenges in multi-file repositories contexts, particularly with function redefinitions. Specifically, if a global function is required for the entire repository, it can be defined in any file, and other files just need to correctly reference it. But they tend to redundantly implement the same function across multiple files, suitable for single-file repositories but erroneous in multi-file scenarios. In C/C++, models incorrectly handle header (.h) and implementation (.cpp) files, leading to redundant declarations and conflicting implementations. These issues highlight a gap in the models' understanding of file-specific roles in programming languages.

\paragraph{File Reference and Linkage Errors}
Correct file referencing is essential in multi-file programming repositories. Models, especially those with larger parameters, generally perform well in establishing basic reference logic, such as using ``import'' in Python and ``\#include'' in C++. However, without review, reference errors are common, likely due to the models' sequential code generation approach, which limits their ability to correct earlier mistakes.
More complex reference issues also arise. Models often struggle to differentiate between global functions and class methods, leading to reference errors when attempting to access class methods directly. These errors are challenging to rectify through review. This indicates a need for models to better grasp the intricacies of programming language structures and conventions.

\subsection{Instruction Following}
\paragraph{Naming Errors}
Proper naming in code is crucial for readability and maintainability. Larger models, like GPT-4-Turbo, while capable of generating syntactically correct code, exhibit deficiencies in this area. It inaccurately modifies function and variable names, disrupting the code's functionality. For instance, in ``Graph BFS DFS'', we instruct the use of the ``top()'' method for stack access, yet GPT-4-Turbo incorrectly labels it as ``getTop()''. Furthermore, the model's lack of attention to plurals and capitalization aggravates these errors. Despite reviews, this issue remains inadequately addressed.

\paragraph{Function Parameters and Overloading Errors}
We observe that even some large models, such as GPT-4-Turbo, neglect the correct number of function parameters, missing critical ones. Function overloading is another similar nuanced aspect that many models mishandle. They often overlook the necessity of multiple constructors or methods with varying parameters. For example, in a task like ``area\_calculation'', models fail to create both parameterized and parameterless constructors, focusing solely on the former. This oversight is not significantly rectified in the review stages, as models mistakenly attribute the errors to the test program. They stubbornly resist correcting their generated code, even when provided with clear instructions.

\paragraph{Type Errors}
Models demonstrate a lack of sensitivity to type conversions, especially in strongly-typed languages like C/C++. Errors in matching const types and misusing pointer types are common, and these missteps are not readily resolved in the review process. However, in weakly-typed languages like Python, such issues are less critical but still present a concern for code accuracy.

\paragraph{Variable Scope and Lifecycle Mismanagement}
Models frequently misuse variables beyond their intended scope or lifecycle. For example, they might attempt to use a loop control variable outside its loop. Another issue is the misunderstanding of private and public members in classes, where models inappropriately access private elements from outside the class. This indicates a gap in the models' understanding of encapsulation and scope management in object-oriented programming.

\subsection{Hallucination}
\paragraph{Fabrication of Variables}
A significant challenge is the models' propensity to fabricate non-existent local variables, a problem typically rectified during review. This issue suggests a fundamental limitation in the models' sequential generation process. Unable to retroactively integrate essential variable definitions, the models end up introducing imaginary variables in the code, resulting in apparent inaccuracies.

\paragraph{Misinterpretation of Data Files}
Models also exhibit a tendency to incorrectly interpret data files as Python libraries. They attempt to import methods from these non-existent libraries, leading to further reference errors. This behavior underscores the intricacy involved in handling file references accurately within code generation tasks.

\subsection{Limitations in Testing}
\label{sec:Limitation in test}

In Acceptance Testing task, we employ an execution-based evaluation method for the generated tests, foregoing manual quality assessments. This approach assumes a test is valid if it can accurately assess standard implementation code. However, we observed that smaller models, such as DeepSeek-Coder-6.7B-Instruct, tend to game this method. They set arbitrary criteria and invariably provide positive feedback, thereby circumventing a genuine evaluation.

Larger models like GPT-4-Turbo fell short of our expectations. They persistently recall and utilize methods in the original repository code, instead of our specially designed versions, leading to frequent import errors. This issue is exemplified in our tests with the ``GeoText'' and ``Stocktrends'' repositories. We modified the original repositories by removing ``\_\_init\_\_.py'' files, expecting models could correctly handle import relationships without them, based on our provided file structures. However, the models continued to follow the import logic of the original repositories, leading to hallucination and inaccurate test generation. This indicates a training data bias, where these models are predisposed to original repository code and show a reluctance to adjust to new circumstance.

To prevent meaningless but executable testing code, structured test templates with explicit instructions and incomplete assertion statements can potentially guide and force models toward meaningful test generation.


\section{Repositories statistics in \devterm}

Table \ref{tab:repo statistics} shows the repository statistics within \devterm.

\label{sec:Repositories statistics}
\begin{table*}[htbp!]
\scalebox{0.8}{
\begin{tabular}{c|c|c|c|c|c|c|c|c}
\toprule
\textbf{Text}                                                                              & \textbf{Language} & \textbf{Domain} & \textbf{\begin{tabular}[c]{@{}c@{}}\#code \\ files\end{tabular}} & \textbf{\begin{tabular}[c]{@{}c@{}}\#code \\ lines\end{tabular}} & \textbf{\begin{tabular}[c]{@{}c@{}}\#code \\ tokens\end{tabular}} & \textbf{\begin{tabular}[c]{@{}c@{}}\#acceptance \\ tests\end{tabular}} & \textbf{\begin{tabular}[c]{@{}c@{}}\#unit \\ tests\end{tabular}} & \textbf{\begin{tabular}[c]{@{}c@{}}Unit test \\ coverage\end{tabular}} \\ \midrule
TextCNN                                                                                    & Python            & DL, NLP         & 5                                                                & 403                                                              & 1566                                                              & 1                                                                      & 10                                                               & 99                                                                     \\ \midrule
ArXiv digest                                                                              & Python            & SE, API         & 1                                                                & 198                                                              & 901                                                               & 4                                                                      & 38                                                               & 94                                                                     \\ \midrule
chakin                                                                                     & Python            & NLP             & 1                                                                & 62                                                               & 225                                                               & 1                                                                      & 1                                                                & 86                                                                     \\ \midrule
readtime                                                                                   & Python            & ALGO            & 3                                                                & 284                                                              & 920                                                               & 4                                                                      & 8                                                                & 95                                                                     \\ \midrule
hone                                                                                       & Python            & SE              & 4                                                                & 274                                                              & 844                                                               & 5                                                                      & 7                                                                & 90                                                                     \\ \midrule
Stocktrends                                                                                & Python            & ALGO            & 1                                                                & 384                                                              & 1350                                                              & 2                                                                      & 7                                                                & 85                                                                     \\ \midrule
GeoText                                                                                    & Python            & NLP             & 2                                                                & 470                                                              & 1701                                                              & 5                                                                      & 4                                                                & 98                                                                     \\ \midrule
lice                                                                                       & Python            & SE              & 2                                                                & 376                                                              & 1329                                                              & 6                                                                      & 25                                                               & 88                                                                     \\ \midrule
PSO                                                                                        & Python            & ALGO            & 2                                                                & 168                                                              & 578                                                               & 1                                                                      & 5                                                                & 93                                                                     \\ \midrule
hybrid images                                                                              & Python            & ALGO, CV        & 1                                                                & 144                                                              & 746                                                               & 1                                                                      & 19                                                               & 90                                                                     \\ \midrule
\begin{tabular}[c]{@{}c@{}}Actor Relationship \\ Game\end{tabular}                                & Java              & ALGO, API            & 8                                                                & 493                                                              & 1453                                                              & 4                                                                      & 16                                                               & 64.32                                                                  \\ \midrule
\begin{tabular}[c]{@{}c@{}}Leftist Trees and \\ Fibonacci Heaps \\ Comparison\end{tabular} & Java              & ALGO            & 3                                                                & 632                                                              & 2009                                                              & 2                                                                      & 2                                                                & 45.32                                                                  \\ \midrule
Redis                                                                                      & Java              & SE, DB          & 9                                                                & 779                                                              & 2546                                                              & 1                                                                      & 17                                                               & 78.6                                                                   \\ \midrule
idcenter                                                                                   & Java              & SE              & 4                                                                & 333                                                              & 1140                                                              & 3                                                                      & 4                                                                & 54.2                                                                   \\ \midrule
\begin{tabular}[c]{@{}c@{}}image \\ similarity\end{tabular}                                & Java              & SE, CV          & 3                                                                & 382                                                              & 1397                                                              & 2                                                                      & 2                                                                & 71.34                                                                  \\ \midrule
xlsx2csv                                                                                   & C/C++             & SE              & 10                                                               & 476                                                              & 1440                                                              & 5                                                                      & 8                                                                & 95.17                                                                  \\ \midrule
\begin{tabular}[c]{@{}c@{}}people\\ management\end{tabular}                                & C/C++             & SE, DB          & 6                                                                & 540                                                              & 2043                                                              & 7                                                                      & 9                                                                & 95.14                                                                  \\ \midrule
\begin{tabular}[c]{@{}c@{}}Area\\ Calculation\end{tabular}                                 & C/C++             & SE              & 7                                                                & 162                                                              & 307                                                               & 3                                                                      & 3                                                                & 90.48                                                                  \\ \midrule
Graph BFS DFS                                                                              & C/C++             & ALGO            & 5                                                                & 667                                                              & 2828                                                              & 5                                                                      & 22                                                               &   /                                                                     \\ \midrule
\begin{tabular}[c]{@{}c@{}}Logistic Management \\ System\end{tabular}                      & C/C++             & SE              & 7                                                                & 630                                                              & 2007                                                              & 7                                                                      & 17                                                               & 99.11                                                                  \\ \midrule
listen-now-frontend                                                                        & JS                & Web             & 6                                                                & 232                                                              & 492                                                               & 1                                                                      & 0                                                                & /                                                                      \\ \midrule
register                                                                                   & JS                & Web             & 6                                                                & 223                                                              & 741                                                               & 3                                                                      & 0                                                                & /                                                                      \\ \bottomrule
\end{tabular}
}
\caption{Repository statistics within \devterm}
\label{tab:repo statistics}
\end{table*}

\end{document}